\documentclass[10pt,twocolumn,letterpaper]{article}

\usepackage{cvpr}
\usepackage{times}
\usepackage{epsfig}
\usepackage{graphicx}
\usepackage{amsmath}
\usepackage{amssymb}
\usepackage{multirow}
\usepackage{galois}
\usepackage{amsthm}
\usepackage{algorithm}
\usepackage{algpseudocode}
\DeclareMathOperator*{\argmax}{argmax}

\usepackage[breaklinks=true,bookmarks=false,colorlinks]{hyperref}

\cvprfinalcopy 

\def\cvprPaperID{630} 

\begin{document}

\title{Crafting a Toolchain for Image Restoration by Deep Reinforcement Learning}


\author{
	Ke Yu $^{1}$ \hspace{9pt} Chao Dong$^{2}$ \hspace{9pt} Liang Lin$^{2, 3}$ \hspace{9pt} Chen Change Loy$^{1}$\\
	$^{1}$\small{CUHK - SenseTime Joint Lab, The Chinese University of Hong Kong}\\
	$^{2}$\small{SenseTime Research} \hspace{13pt} $^{3}$\small{Sun Yat-sen University}\\
	{\tt\small \{yk017, ccloy\}@ie.cuhk.edu.hk \hspace{5pt} \{dongchao, linliang\}@sensetime.com}
}

\maketitle

\begin{abstract}
	We investigate a novel approach for image restoration by reinforcement learning. Unlike existing studies that mostly train a single large network for a specialized task, we prepare a toolbox consisting of small-scale convolutional networks of different complexities and specialized in different tasks. Our method, RL-Restore, then learns a policy to select appropriate tools from the toolbox to progressively restore the quality of a corrupted image. 
	We formulate a step-wise reward function proportional to how well the image is restored at each step to learn the action policy. We also devise a joint learning scheme to train the agent and tools for better performance in handling uncertainty.
	In comparison to conventional human-designed networks, RL-Restore is capable of restoring images corrupted with complex and unknown distortions in a more parameter-efficient manner using the dynamically formed toolchain\footnote{Codes and data are available at \url{http://mmlab.ie.cuhk.edu.hk/projects/RL-Restore/}}.
	\vspace{-0.4cm}
\end{abstract}

\section{Introduction}

Deep convolutional neural network (CNN) has achieved immense success, not only in high-level vision tasks, but also low-level vision tasks such as deblurring \cite{nah2017deep, sun2015learning, xu2014inverse}, denoising \cite{chen2015learning, lefkimmiatis2016non}, JPEG artifacts reduction \cite{dong2015compression, guo2016building, wang2016d3} and super-resolution \cite{dong2016image, kim2016accurate, hui2016depth, tai2017memnet,wang2018recovering}.
In particular, good performance and fast testing speed are demonstrated over conventional model-based optimization methods.

Owing to the discriminative nature of CNN, most of these models are trained to handle a specialized low-level vision task. 
In JPEG artifacts reduction~\cite{dong2015compression}, for instance, different networks for different compression qualities have been designed to achieve satisfactory restoration.
In the case of super-resolution~\cite{dong2016image}, it is common to have different networks to handle different scaling factors.
Some recent studies~\cite{guo2016one, tai2017memnet} have shown the possibility of handling multiple distortion types or coping with different levels of degradation at once using CNN. Nevertheless, this usually comes with the expenses of using much deeper networks. In addition, such networks process all images with the same structure, despite some of which are inherently less difficult and can be restored in a cheaper way.

In this paper, we explore the possibility of having some smaller-scale but specialized CNNs to solve a harder restoration task collaboratively.
Our idea departs from the current philosophy that one would need a large-capacity CNN to solve a complex restoration task. Instead, we wish to have a set of tools (based on small CNNs) and learn to use them adaptively for solving the task at hand.
The aforementioned idea could provide new insights how CNN can be used for solving real-world restoration tasks, of which images are potentially contaminated with a mix of distortions, \eg, blurring, noise and blockiness after several stages of processing. 
Moreover, the new approach may lead to parameter-efficient restoration in comparison to existing CNN-based models.
In particular, tools of different complexities can be selected based on the severity of distortion.

Towards this goal, we present a framework that treats image restoration as a decision making process by which an agent would adaptively select a sequence of tools to progressively refine an image, and the agent may choose to stop if the restored quality is deemed satisfactory. 
In our framework, we prepare a number of light-weight CNNs with different complexities. They are task-specific aiming to handle different types of restoration assignments including deblurring, denoising, or JPEG artifacts reduction.
Choosing the order of tools is formulated in a reinforcement learning (RL) framework.
An agent learns to decide the next best tool to select by analyzing the content of the restored image in the current step and observing the last action chosen. Rewards are accumulated when the agent improves the quality of the input image.

\begin{figure}[t]
	\begin{center}
		\includegraphics[width=\linewidth]{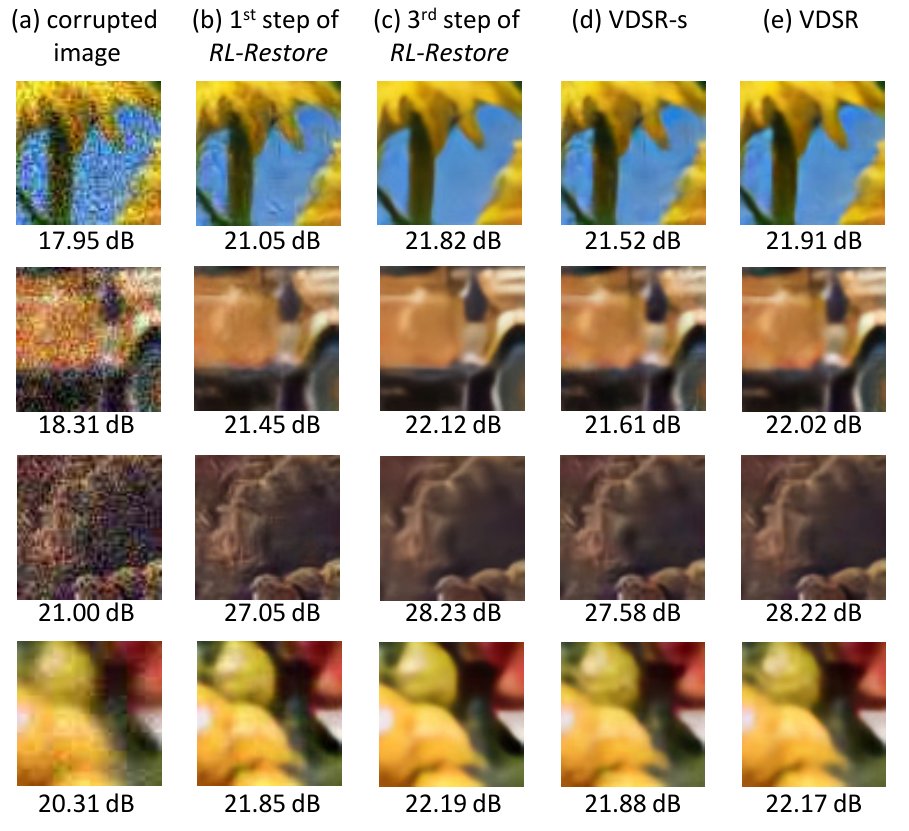}
		\vskip -0.1cm
		\caption{(a) shows images corrupted by complex distortions. (b-c) depict some chosen steps of the decision process to restore an image by \textit{RL-Restore}. At each step, a specific tool is selected by the agent to improve the image quality. (d-e) are CNN-based results, where (d) has comparable parameters to \textit{RL-Restore} while (e) has twice more. PSNR values are presented for better comparison.
		}
		\label{fig:overview}
	\end{center}
	\vskip -0.4cm
\end{figure}

We refer to the proposed framework as \textit{RL-Restore}. We summarize our \textbf{contributions} as follows:

\noindent 1) We present a new attempt to address image restoration in a reinforcement learning framework. Unlike existing methods that deploy a single and potentially large network structure, \textit{RL-Restore} enjoys the flexibility of using tools of different capacities to achieve the desired restoration.

\noindent 2) We propose a joint learning scheme to train the agent and tools simultaneously so that the framework possesses better capability in coping with new and unknown artifacts emerged in the mid of processing.  

\noindent 3) We show that the dynamically formed toolchain performs competitively against strong human-designed networks with less computational complexity. Our approach can cope with unseen distortions to certain extent. Interestingly, our approach is more transparent than existing methods as it can reveal how complicated distortions could be removed step by step using different tools.

Figure \ref{fig:overview}(b-c) illustrate a learned policy to restore an image corrupted by multiple distortions, where image quality is refined step-by-step. The results of two baseline CNN models are depicted in Figure \ref{fig:overview}(d-e), where (d) has similar number of parameters as ours (agent $+$ tools applied),  while (e) has twice more. As we will further present in the experimental section, \textit{RL-Restore} is superior to CNN approaches given similar complexity and it requires 82.2\% fewer computations to achieve the same performance as a single large CNN.

\section{Related Work}

\noindent \textbf{CNN for Image Restoration}.
Image restoration is an extensively studied topic that aims at estimating the clear/original image from a corrupted/noisy observation. Convolutional neural networks (CNN) based methods have demonstrated outstanding performance in various image restoration tasks. 
Most of these studies train a single network specializing on the task at hand, \eg, deblurring \cite{nah2017deep, sun2015learning, xu2014inverse}, denoising \cite{chen2015learning, lefkimmiatis2016non}, JPEG artifacts reduction \cite{dong2015compression, guo2016building, wang2016d3} and super-resolution \cite{dong2016image, hui2016depth, kim2016accurate, kim2016deeply, lai2017deep, tai2017image, tai2017memnet,wang2018recovering}.
Our work offers an alternative that is more parameter efficient yet adaptive to the form of distortions. 

There are several pioneering studies that deal with multiple degradations simultaneously. By developing a 20-layer deep CNN, Kim~\etal~\cite{kim2016accurate} use a single model to handle multi-scale image super-resolution. Guo~\etal~\cite{guo2016one} build a one-to-many network that can handle images with different levels of compression artifacts. 
Zhang~\etal~\cite{zhang2017beyond} propose a 20-layer deep CNN to address multiple restoration tasks simultaneously, including image denoising, JPEG artifacts reduction and super-resolution. None of these studies considers mixed distortion, where a single image is affected by multiple distortions. 
Different from the aforementioned works, we are interested to explore if smaller-scale CNNs of 3 to 8 layers could be used to jointly restore images that are contaminated with mixed distortions.

There exist approaches~\cite{chen2015compressing,han2015deep,hinton2015distilling} that can be used to compress a large network to a smaller one for computational efficiency. In the domain of image restoration, recursive neural networks \cite{kim2016deeply,tai2017image,tai2017memnet} are investigated to reduce network parameters. However, the computational cost is still high due to the large number of recursions.
The objective of our work is orthogonal to the aforementioned studies -- our framework saves parameters and computation through learning a policy to make decision in selecting appropriate CNNs for a task rather than compressing an existing one. 

\noindent \textbf{Deep Reinforcement Learning}.
Reinforcement learning is a powerful tool for learning an agent making sequential decisions to maximize accumulative rewards. Early works of RL mainly focus on robotic control \cite{lin1993reinforcement, vermorel2005multi}. Recently traditional RL algorithms are incorporated in deep learning frameworks and are successfully applied in various domains such as game agents \cite{lillicrap2015continuous, mnih2015human, silver2016mastering, silver2017mastering} and neural network architecture design \cite{baker2016designing, zoph2016neural}. Attention is also drawn to deep RL in the field of computer vision \cite{ba2014multiple, cao2017attention, he2018merge, huang2017learning, liang2017deep, liu2016learning, mnih2014recurrent, ren2017deep, yoo2017action}. 
%
%
For instance, Huang~\etal~\cite{huang2017learning} use RL to learn an early decision policy for speeding up object tracking by CNN.
Cao~\etal~\cite{cao2017attention} explore deep RL algorithms in low-level vision and apply attention mechanism \cite{mnih2014recurrent} to face hallucination. 
In this study, we investigate restoration tool selection in a RL framework. The problem is new in the literature.


\section{Learning a Restoration Toolchain}
\label{sec:methodology}

\begin{figure}[t]
	\begin{center}
		\includegraphics[width=\linewidth]{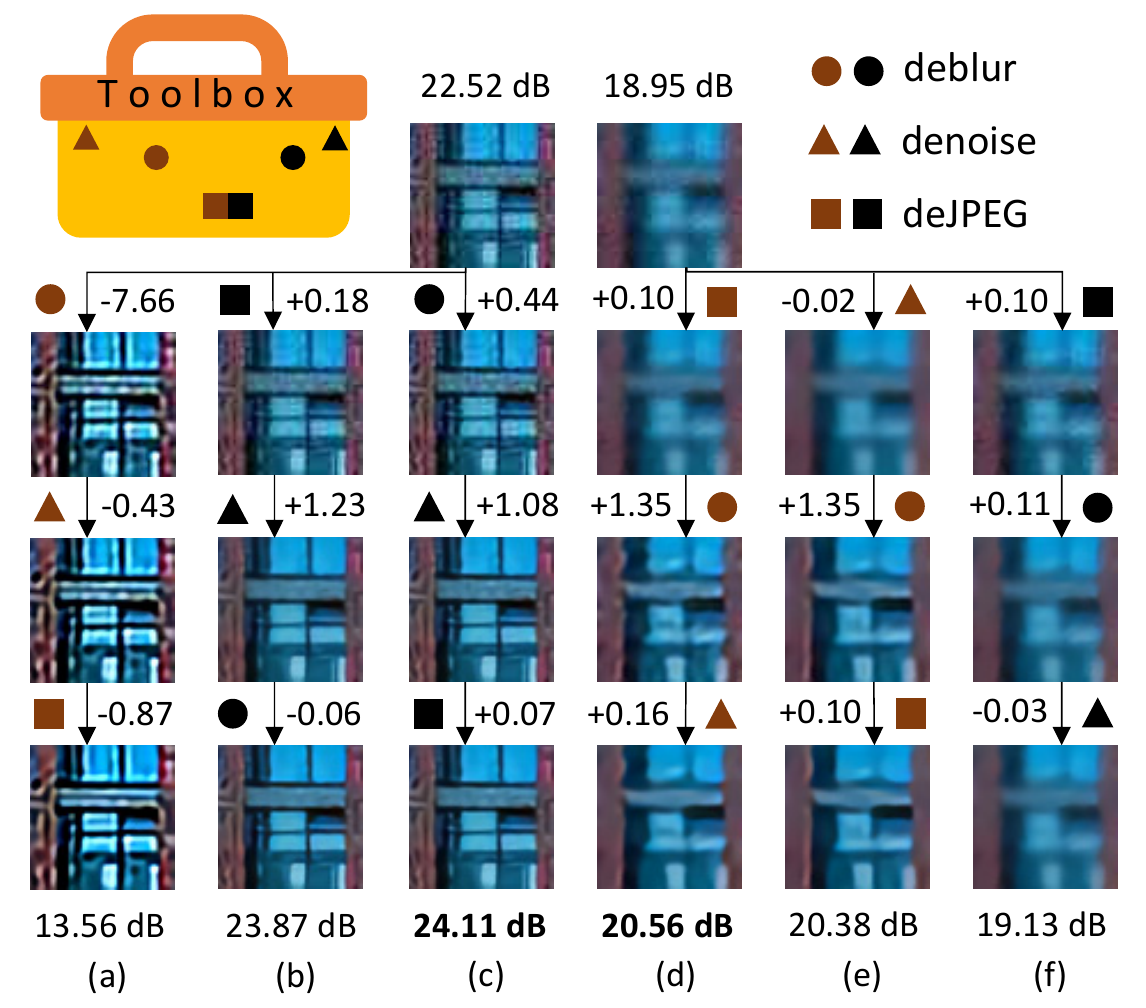}
		\vskip -0.1cm
		\caption{\textbf{Different toolchains for image restoration}. We perform a preliminary test here. Given two distorted images and the corresponding appropriate toolchains as (c) and (d), we construct other toolchains by rearranging the order (represented by shape) or adjusting the level (represented by color) of the selected tools. The restored results indicate that such minor changes of a toolchain could lead to very different performance.}
		\label{fig:problem_analysis}
	\end{center}
	\vspace{-0.6cm}
\end{figure}

\begin{figure*}[t]
	\centering
	\includegraphics[width=0.9\textwidth]{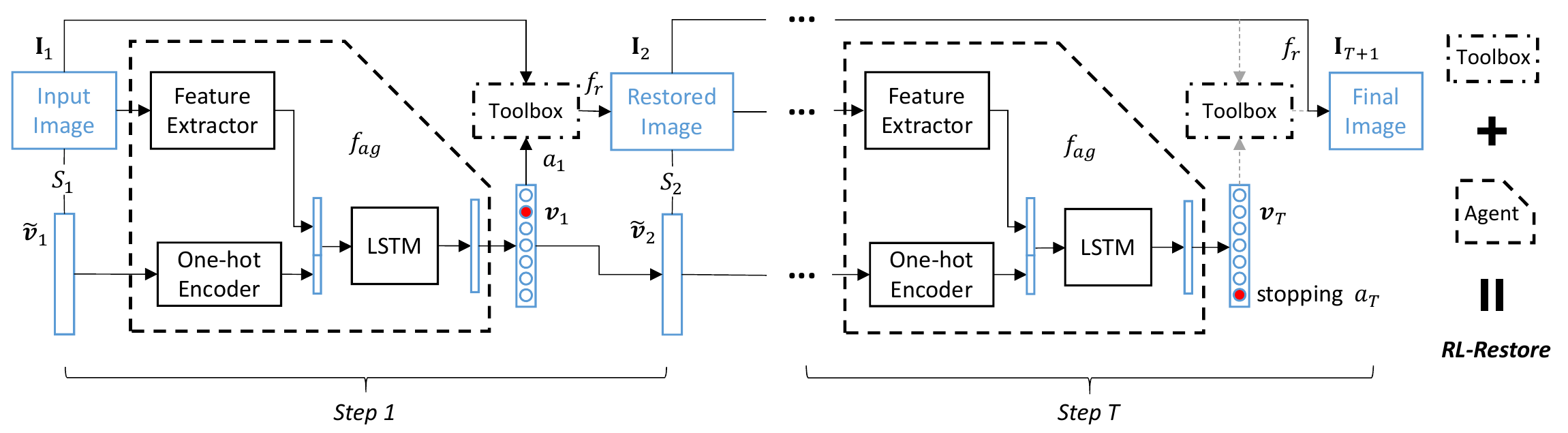}
	\caption{\textbf{Illustration of our \textit{RL-Restore} framework.} At each step $t$, the agent $f_{ag}$ observes the current state $S_{t}$, including the current restored image $\mathbf{I}_{t}$ and input value vector $\boldsymbol{\tilde{v}}_{t}$, which is the output of the agent at the previous step. Note that $\mathbf{I}_{1}$ represents the input image and $\boldsymbol{\tilde{v}}_{1}$ is a zero vector. Based on the maximum value of the agent's output $\boldsymbol{v}_{t}$, an action $a_t$ is selected and the corresponding tool is used to restore the current image. After restoration process $f_r$, with the newly restored image $\mathbf{I}_{t+1}$ and value vector $\boldsymbol{\tilde{v}}_{t+1}=\boldsymbol{v}_{t}$, \textit{RL-Restore} conducts another step of restoration iteratively until the stopping action is selected.}
	\label{fig:framework}
	\vskip -0.3cm
\end{figure*}

\noindent{\textbf{Problem Definition.}} Given a distorted image $\mathbf{I}_{dis}$, our goal is to restore a clear image $\mathbf{I}_{res}$ that is close to the ground truth image $\mathbf{I}_{gt}$. The distortion process can be formulated as:
\begin{equation}
\begin{aligned}
\mathbf{I}_{dis} = D(\mathbf{I}_{gt});\quad D = D_n~\comp~\cdots~\comp~D_1,
\end{aligned}
\end{equation}
where $\comp$ denotes function composition and each of $D_1,\dots,D_n$ represents a specific type of distortion. In contrast to existing methods \cite{chen2015learning,dong2016image,nah2017deep,tai2017memnet,wang2016d3} that concentrate on a single type of distortion, we intend to handle a mix of multiple distortions (\ie,~$n>1$). For example, the final output image may be sequentially affected by out-of-focus blur, exposure noise and JPEG compression. In such a case, the number of distortions $n$ is 3, and $D_1$, $D_2$, $D_3$ represent blur, noise and compression, respectively. To address mixed distortions, we propose to restore the corrupted image step by step with a sequence of restoration tools.

\noindent{\textbf{Challenges.}} 
The task of tool selection is non-trivial and presents unique challenges to RL. First, \textit{the choice of the restoration type, level and the processing order all influence the final performance}. An example is shown in Figure~\ref{fig:problem_analysis}, where the images are corrupted by two different combinations of distortions. 
With an appropriate toolchain, as in Figure~\ref{fig:problem_analysis} (c, d), the image quality and the Peak Signal-to-Noise Ratio (PSNR) values are improved sequentially. Then we slightly re-arrange the tools order as in Figure~\ref{fig:problem_analysis}(b, e) or adjust the restoration level of the tools as in Figure~\ref{fig:problem_analysis}(a, f). The results indicate that minor changes in a toolchain can severely impact the restoration performance. Specifically, using improper tools may lead to unnatural outputs, such as over-sharpening in Figure~\ref{fig:problem_analysis}(a) and blurring in Figure~\ref{fig:problem_analysis}(f). Even the tools are well chosen, an inappropriate order could decrease the performance (Figure~\ref{fig:problem_analysis}(b, e)). Since the sequence of toolchain dramatically influences the results, selecting which tool to use at each step becomes crucial.

When the tools are trained on specific tasks, we encounter another problem that  \textit{none of the tools can perfectly handle the `middle state'}, which refers to the intermediate result after several steps of processing. As most distortions are irreversible, the restoration of their mixture is not a simple composition of the corresponding restorers. New  artifacts could be introduced in the middle states. For example, the deblurring operation will also enhance the noises, causing the following denoisers fail in removing the newly introduced artifacts. The challenge is unique to our task. 

To address the first challenge, we treat the sequential tool selection problem as a Markov Decision Process (MDP) and solve it in a deep reinforcement learning manner. 
To address the second challenge, we propose a training scheme to refine the agent and tools jointly so that the tools are more well-informed with the middle states observable by the agent. We first provide an overview of the proposed framework as follows.


\noindent{\textbf{Overview of \textit{RL-Restore}.}}
The proposed framework aims at discovering a toolchain given a corrupted input image. As shown in Figure~\ref{fig:framework}, \textit{RL-Restore} consists of two components: 1) a toolbox that contains various tools for image restoration and 2) an agent with a recurrent structure that dynamically chooses a tool at each step or an early stopping action. We cast the tool selection process as a reinforcement learning procedure -- a sequence of decision on tool selection is made to maximize a reward proportional to the quality of the restored image.
Next, we first describe a plausible setting of toolbox and then explain the details of the agent.

\subsection{Toolbox}
\label{subsec:toolbox}

The toolbox contains a set of tools that might be applied to the corrupted image. Our goal is to design a powerful and light-weight toolbox, we thus restrict each tool to be proficient in a specific task. That is, each tool is trained only on a narrow range of distortions. To further reduce the overall complexity, we use smaller networks for easier tasks. For the purpose of our research, we prepare 12 tools as shown in Table~\ref{table:toolbox}, where each tool is assigned to address a certain level of Gaussian blur, Gaussian noise or JPEG compression. We apply a three-layer CNN (as in \cite{dong2016image}) for slight distortions and a deeper eight-layer CNN for severe distortions. 
%
Note that the tools need not be restricted to solve the aforementioned distortions. We made these selections since they are typically considered in the literature of image restoration. In practice, one could design their tools with appropriate complexity based on the task at hand.

As discussed at the beginning of Sec.~\ref{sec:methodology}, a finite set of tools is not perfect to handle new artifacts emerged in middle states. To address this issue, we propose two strategies : 1) To increase robustness of the tools, we add slight Gaussian noises and JPEG compression to all the training data. 2) After training the agent, all tools are jointly fine-tuned on the basis of the well-trained toolchains. Then the tools will be more adaptive to the agent task, and be able to deal with middle states more robustly. We discuss the training steps in Sec.~\ref{subsec:training}. Experiments in Sec.~\ref{sec:exp} validate the effectiveness of the proposed strategies.

\begin{table}[t] \centering \small
	\center
	\caption{Tools in the toolbox. We consider three types of distortion and various degradation levels. Each tool is either a 3-layer CNN or an 8-layer CNN according to the distortion it targets to solve.} \vspace{0.05cm}
	\begin{tabular}{c|c|c}
		\hline
		Distortion Type & \multirow{2}*{Distortion Level Interval}  & CNN   \\
		(Parameters)    & & Depth \\
		\hline\hline
		\multirow{2}*{Gaussian Blur ($\sigma$)} & [0, 1.25], [1.25, 2.5]& 3 \\
		\cline{2-3}
		& [2.5, 3.75], [3.75, 5] & 8 \\
		\hline
		\multirow{2}*{Gaussian Noise ($\sigma$)} & [0, 12.5], [12.5, 25]  & 3 \\
		\cline{2-3}
		& [25, 37.5], [37.5, 50] & 8 \\
		\hline
		\multirow{2}*{JPEG Compression (Q)} & [60, 100], [35, 60] & 3 \\ 
		\cline{2-3}
		& [20, 35], [10, 20]& 8 \\
		\hline
		
	\end{tabular}
	\label{table:toolbox}
\end{table}

\subsection{Agent}
The processing pipeline of \textit{RL-Restore} is shown in Figure \ref{fig:framework}.
Given an input image, the agent first selects a tool from the toolbox and uses it to restore the image, then the agent chooses another tool according to the previous result and repeats the restoration process until it decides to stop. We will first clarify some terminologies such as action, state and reward, and then go into the details of the agent structure and restoration procedure.

\vspace{0.1cm}
\noindent{\textbf{Action.}} 
The action space, denoted as $A$, is a set of all possible actions that the agent could take. At each step $t$, an action $a_t$ is selected and applied to the current input image. Each action represents a tool in the toolbox and there is one additional action that represents stopping. If there are $N$ tools in the toolbox, then the cardinality of $A$ is $N+1$. Hence, the output, $\boldsymbol{v}_{t}$, of the agent is an $(N+1)$-dimensional vector that implicates the value of each action. 
Once the stopping action is chosen, the restoration procedure will be terminated and the current input image will become the final result.  

\vspace{0.1cm}
\noindent{\textbf{State.}} 
The state contains information that the agent could observe. In our formulation, the state is formulated as $S_t=\{\mathbf{I}_{t},\boldsymbol{\tilde{v}}_t\}$, where $\mathbf{I}_{t}$ is the current input image, and $\boldsymbol{\tilde{v}}_t$ is the past historical action vector. At step 1, $\mathbf{I}_{1}$ is the input image and $\boldsymbol{\tilde{v}}_{1}$ is a zero vector. 
The state provides rich contextual knowledge to the agent. 1) The current input image $\mathbf{I}_{t}$ is essential because the selected action will be directly applied to this image to derive a better restored result. 2) the information of previous action vector $\boldsymbol{\tilde{v}}_t$, which is the output value vector of the agent at $t-1$ step, \ie, $\boldsymbol{\tilde{v}}_t=\boldsymbol{v}_{t-1}$, is important too. The knowledge of the previous decision could help the action selection at the current step. 
This is found to work better empirically than using $\mathbf{I}_{t}$ only.

\vspace{0.1cm}
\noindent{\textbf{Reward.}} 
The reward drives the training of the agent as it learns to maximize the cumulative reward. The agent is supposed to learn a good policy so that the final restored image is satisfactory. We wish to ensure that the image quality is enhanced at each step, therefore a stepwise reward is designed as follows:
\vspace{-0.1cm}
\begin{equation}
r_t = P_{t+1}-P_t,
\end{equation}
where $r_t$ is the reward function at step $t$, $P_{t+1}$ denotes the PSNR between $\mathbf{I}_{t+1}$ and the reference image $\mathbf{I}_{gt}$ at the end of the $t$-th step restoration, and $P_t$ represents the input PSNR at step $t$. The cumulative reward can be written as $R=\sum_{t=1}^{T}r_t = P_{T+1}-P_1$, which is the overall PSNR gain during the restoration procedure, and it is maximized to achieve optimal enhancement. Note that it is flexible to use other image quality metrics (\eg,~perceptual loss~\cite{johnson2016perceptual}, GAN loss~\cite{ledig2017photo}) as the reward in our framework. The investigation is beyond the focus of this paper. 

\vspace{0.1cm}
\noindent{\textbf{Structure.}} 
At each step $t$, the agent assesses the value of each action given the input state $S_{t}$, which can be formulated as follows:
\begin{equation}
\begin{aligned}
\boldsymbol{v}_t=f_{ag}(S_{t}; W_{ag}),
\end{aligned}
\end{equation}
where $f_{ag}$ indicates the agent network and $W_{ag}$ denotes its parameters. The vector $\boldsymbol{v}_t$ represents the value of actions. The action with the maximum value is selected as $a_t$, \ie, $a_t = {\argmax}_a {v}_{t,a}$, where $v_{t,a}$ indicates the element of value vector $\boldsymbol{v}_t$ corresponding to action $a$.

The agent is composed of three modules as depicted in Figure \ref{fig:framework}. The first module, named feature extractor, is a four-layer CNN followed by a fully-connected (fc) layer that outputs a 32-dimensional feature. The second module is a one-hot encoder with $N+1$ dimensional input and $N$ dimensional output, preserving the information of the previous chosen action. Note that the output is one dimension lower than the input, because the stopping action cannot be adopted at the previous step, and thus we simply drop the last dimension. The outputs of the first two modules are concatenated into the input of the third module, which is a Long Short-Term Memory (LSTM) \cite{hochreiter1997long}. The LSTM not only observes the input state, but also stores historical states in its memory, which offers contextual information of historical restored images and actions. Finally, with another fc layer following LSTM, a value vector $\boldsymbol{v}_{t}$ is derived for tool selection. 

\vspace{0.1cm}
\noindent{\textbf{Restoration.}}
Once an action $a_t$ is obtained based on the maximum value in $\boldsymbol{v}_{t}$, the corresponding tool will be applied to the input image $\mathbf{I}_{t}$ to get a new restored image:
\begin{equation}
\begin{aligned}
\mathbf{I}_{t+1}= f_r(\mathbf{I}_{t},a_t;W_r),
\end{aligned}
\end{equation}
where $f_r$ denotes the restoration fucntion and $W_r$ indicates the parameters of a tool in the toolbox. If a stopping action is selected, $f_r$ represents an identity mapping.
By denoting $\mathbf{I}_{dis}$ and $\mathbf{I}_{res}$ as the input distorted image and final restored output respectively, the overall procedure of restoration can be expressed as:
\begin{equation}
\begin{cases}
\mathbf{I}_1 = \mathbf{I}_{dis}         & \\
\mathbf{I}_{t+1} = f(\mathbf{I}_{t};W)  & 1 \leq t \leq T \\
\mathbf{I}_{res} = \mathbf{I}_{T+1},    &
\end{cases}
\end{equation}
where $f = [f_{ag};f_r]$ and $W=[W_{ag};W_r]$. $T$ is the step when the stopping action is chosen. We also set a maximum step $T_\mathrm{max}$ to prevent excessive restoration. When $t=T_\mathrm{max}$ and the stopping action is not selected, we will terminate the restoration process after the current step. In other words, we add a constraint that $T\leq{T_\mathrm{max}}$.

\subsection{Training}
\label{subsec:training}

The training of tools follows a standard setting in \cite{kim2016accurate}, where a mean square error (MSE) $\frac{1}{2}\|\mathbf{y}-h(\mathbf{x})\|_2^2$ is minimized. The ground truth image, input image and the tool are denoted as $\mathbf{y},\mathbf{x}$ and $h$, respectively. As for the agent, the training is addressed by deep Q-learning \cite{mnih2015human} since we do not have \textit{a priori} knowledge about the correct action to choose. 
In the proposed framework, each element of $\boldsymbol{v}_t$ is an action value as defined in \cite{mnih2015human}, so the loss function can be written as $L=(y_t-v_{t, a_t})^2$ where
\begin{equation}
y_t = \begin{cases}
r_t + \gamma\max_{a'}v_{t+1,a'}  & 1 \leq t < T \\
r_T                        & t = T,
\end{cases}
\end{equation}
and $\gamma=0.99$ is a discount factor. We also employ a target network $f'_{ag}$ to stabilize training, which is a clone of $f_{ag}$ and updates its parameters every $C$ steps while training. In the above formula, $v_{t+1,a'}$ is derived from $f'_{ag}$ and $v_{t, a_t}$ is from $f_{ag}$. 
While training, episodes are randomly selected from a replay memory, and there are two updating strategies as proposed in \cite{hausknecht2015deep}, where `random updates' refer to updating from a random point of each episode and proceeding a fixed number of steps, and `sequential updates' indicate that all the updates begin at the beginning of the episode and proceed to its endpoint. In \cite{hausknecht2015deep}, it is claimed that both updating strategies have similar performance. Since our toolchain is not too long, we simply adopt `sequential updates' where each training sequence contains an entire toolchain. 

\vspace{0.1cm}
\noindent \textbf{Joint Training}.
As discussed in Section~\ref{subsec:toolbox}, none of the tools can perfectly handle the middle state, where new and complex artifacts may be introduced in the previous steps of restoration. In order to address this issue, we propose a joint training algorithm, as shown in Algorithm~\ref{alg:joint}, to train the tools in an end-to-end manner so that all the tools can learn to deal with the middle state. Specifically, for each toolchain in a batch, the distorted image $\mathbf{I}_1$ is forwarded to get a restored result $\mathbf{I}_{T+1}$. Given a final MSE loss, the gradients then pass backward along the same toolchain. Meanwhile, the gradients of each tool are accumulated within a batch, and finally an average of gradient is used to update the corresponding tool. The above updating process is repeatedly conducted for a few iterations.

\begin{algorithm}[t]
	\caption{Joint training algorithm (1 iteration)}
	\label{alg:joint}
	\small
	\begin{algorithmic}
		\State Initialize counters $c_1,c_2,\dots,c_N=0$
		\State Initialize gradients $G_{1},G_{2},\dots,{G_N}=0$
		\For{$m=1,M$}\Comment{For each toolchain}
		\State $\mathbf{I}_1 \gets$ Input image
		\For{$t=1,T$}\Comment{Forward paths}
		\State $a_t \gets f_{ag}(S_t)$
		\State $\mathbf{I}_{t+1} \gets f_r(\mathbf{I}_t,a_t)$ 
		\EndFor
		\State $L \gets \frac{1}{2}\|\mathbf{I}_{gt}-\mathbf{I}_{T+1}\|_2^2$
		\For{$t=T$ \textbf{to} $1$ \textbf{step} $-1$}\Comment{Backward paths}
		\State $c_{a_t} \gets c_{a_t} + 1$
		\State $G_{a_t} \gets G_{a_t} + \partial{L} / \partial{W_{a_t}}$
		\State $L \gets \mathbf{I}_t \cdot \partial{L} / \partial{\mathbf{I}_{t}}$
		\EndFor
		\EndFor
		\For{$i=1,N$}\Comment{Update tools}
		\If{$c_i>0$}
		\State $W_i \gets W_i - \alpha{G_i}/c_i$
		\EndIf
		\EndFor
	\end{algorithmic}
\end{algorithm}

\vspace{0.1cm}
\noindent \textbf{Implementation Details}.
In our implementation, the training of tools is similar to \cite{kim2016accurate}, where all experiments run over $80$ epochs ($3.2\times10^5$ iterations) with a batch size of $64$. The initial learning rate is $0.1$ and it decreases by a factor of $0.1$ every $20$ epochs. For joint training, we set $M=64,\alpha=0.0001$ in Algorithm~\ref{alg:joint}, denoting the batch size and learning rate respectively. The joint training runs over $2\times10^5$ iterations.
While training the agent, we use Adam \cite{kingma2014adam} optimizer and a batch size of $32$. The maximum step $T_{\mathrm{max}}$ is set to be $3$ empirically and the size of replay memory is chosen as $5\times10^5$. The updating frequency $C=2,500$ so that the target network $f'_{ag}$ is copied from the latest agent network $f_{ag}$ every $2,500$ iterations. The learning rate is decayed exponentially from $2.5\times10^{-4}$ to $2.5\times10^{-5}$ within $5\times10^5$ iterations. 

\section{Experiments}
\label{sec:exp}

\noindent{\textbf{Datasets and Evaluation Metrics.}}
We perform experiments on the DIV2K dataset~\cite{Agustsson_2017_CVPR_Workshops}, which is the most recent large-scale and high-quality dataset for image restoration. The 800 DIV2K training images are divided into two parts: 1) the first 750 images for training and 2) the rest 50 images for testing. The DIV2K validation images are used for validation. Training images are augmented by down-scaling with factors of 2, 3 and 4. The images are then cropped into 63$\times$63 sub-images, forming our training set and testing set with 249,344 and 3,584 sub-images, respectively.

\begin{figure}[t]
	\vskip -0.15cm
	\begin{center}
		\includegraphics[width=\linewidth]{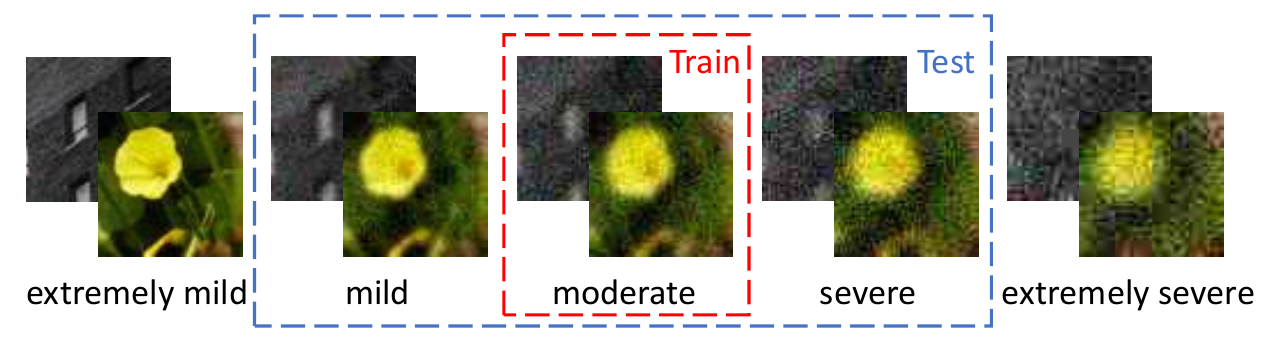}
		\caption{Different levels of distortions.}
		\label{fig:mild_to_severe}
	\end{center}
	\vskip -0.35cm
\end{figure}

We employ mixed distortions for agent training and testing. Specifically, a sequence of Gaussian blur, Gaussian noise and JPEG compression is added to the training images with random levels. The standard deviations of Gaussian blur and Gaussian noise are uniformly distributed in [0, 5] and [0, 50], respectively, while the quality of JPEG compression is subjected to a uniform distribution in [10, 100]. All mixed distortions are categorized into five groups, as shown in Figure~\ref{fig:mild_to_severe}, from extremely mild to extremely severe. We discard two extreme cases that are either too easy or too hard for restoration. Training and testing are performed on the moderate group. To further test the generalization ablity, we also perform testing on mild and severe groups that are not included in the training data. 


\noindent{\textbf{Comparisons.}}
We compare \textit{RL-Restore} with DnCNN~\cite{zhang2017beyond} and VDSR~\cite{kim2016accurate}, which are the state-of-the-art models for image restoration and super-resolution, and both of them are capable of handling multiple degradations. DnCNN and VDSR share similar structure with 20 convolutional layers while batch normalization is adopted in DnCNN. Their parameters are over 0.6 million (shown in Table \ref{table:complexity}). In contrast, the complexity of  \textit{RL-Restore} (including the agent and the selected tools\footnote{The complexity of toolchain is calculated under the assumption that each tool is chosen with equal probabilities and the stopping action is ignored. We do not adopt batch normalization in any model.}) is only about a third of those for DnCNN and VDSR, with 0.19 million parameters in total. A much larger gap can be observed on computations when we refer to the number of multiplications on a $63 \times 63$ input image. 
For a fair comparison with \textit{RL-Restore}, we shrink VDSR from 20 to 15 layers (42 filters in each layer) to form a new baseline, named VDSR-s, which bares similar complexity as \textit{RL-Restore}.
Following the same training strategy in \cite{kim2016accurate, zhang2017beyond}, we first train the baselines with the agent training set. Then we fine-tune the models with both the agent and tools training sets till convergence.

\begin{table}[t] \centering \footnotesize
	\center
	\caption{Complexity of baselines and \textit{RL-Restore}.} 
	\vskip -0.25cm
	\begin{tabular}{c|c|c|c|c}
		\hline
		Model						& DnCNN & VDSR & VDSR-s & \textit{RL-Restore} \\ \hline \hline
		Parameters ($\times10^5$)   & 6.69  & 6.67 & 2.09   & 1.96                \\ \hline
		Computations ($\times10^9$) & 2.66  & 2.65 & 0.828  & 0.474               \\ 
		\hline
	\end{tabular}
	\label{table:complexity}
	\vskip -0.2cm
\end{table}

\subsection{Quantitative Evaluation on Synthetic Dataset}

\begin{table}[t] \centering \footnotesize
	\centering
	\caption{Quantitative results on DIV2K test sets.}
	\begin{tabular}{c|c|c|c|c|c|c}
		\hline
		Test Set    & \multicolumn{2}{c|}{Mild (unseen)}  & \multicolumn{2}{c|}{Moderate} & \multicolumn{2}{c}{Severe (unseen)} \\ \hline
		Metric      & PSNR  & SSIM & PSNR  & SSIM & PSNR & SSIM  \\\hline \hline
		DnCNN       & 28.03          & \textbf{0.6503} & 26.42 & 0.5554 & 24.99 & 0.4658 \\ 
		VDSR        & \textbf{28.04} & 0.6496 & 26.40 & 0.5544 & 24.90 & 0.4629 \\ 
		VDSR-s      &  27.69        & 0.6383 & 25.99 & 0.5399 & 24.50 & 0.4505 \\ 
		\textit{RL-Restore} & 28.04  & 0.6498 & \textbf{26.45} & \textbf{0.5587} & \textbf{25.20} & \textbf{0.4777} \\ 
		\hline
	\end{tabular}
	\label{table:synthetic}
	\vskip -0.25cm
\end{table}

\begin{figure*}[t]
	\begin{center}
		\includegraphics[width=\linewidth]{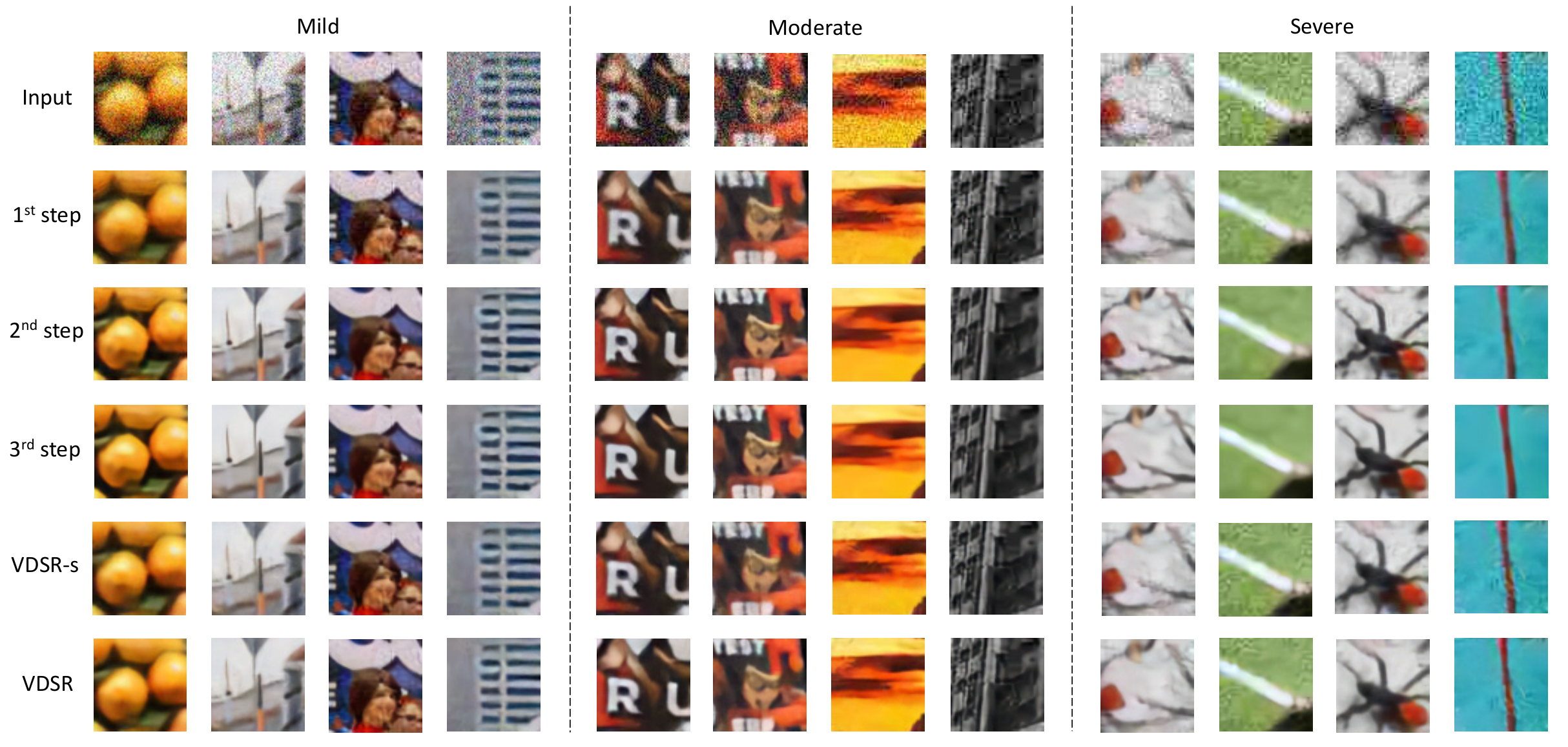}
		\vskip -0.15cm
		\caption{Qualitative comparisons with baselines on synthetic dataset.}
		\label{fig:synthetic_comparison}
	\end{center}
	\vspace{-0.65cm}
\end{figure*}

We present quantitative results of \textit{RL-Restore} and baselines on different test sets in Table~\ref{table:synthetic}. The results on mild and moderate sets show that our approach is apparently superior to VDSR-s while comparable to DnCNN and VDSR, demonstrating that the proposed \textit{RL-Restore} could achieve the same performance as a deep CNN with much lower complexity. It is worth noting that on severe test set \textit{RL-Restore} surpasses DnCNN and VDSR by 0.2 dB and 0.3 dB, respectively, where the distortions are not observed in the training data. It indicates that our RL-based approach is more flexible in handling unseen distortions, while it is more difficult for a fixed CNN to generalize towards unseen cases. Visual results are shown in Figure~\ref{fig:synthetic_comparison}.

\begin{figure}[t]
	\begin{center}
		\includegraphics[width=\linewidth]{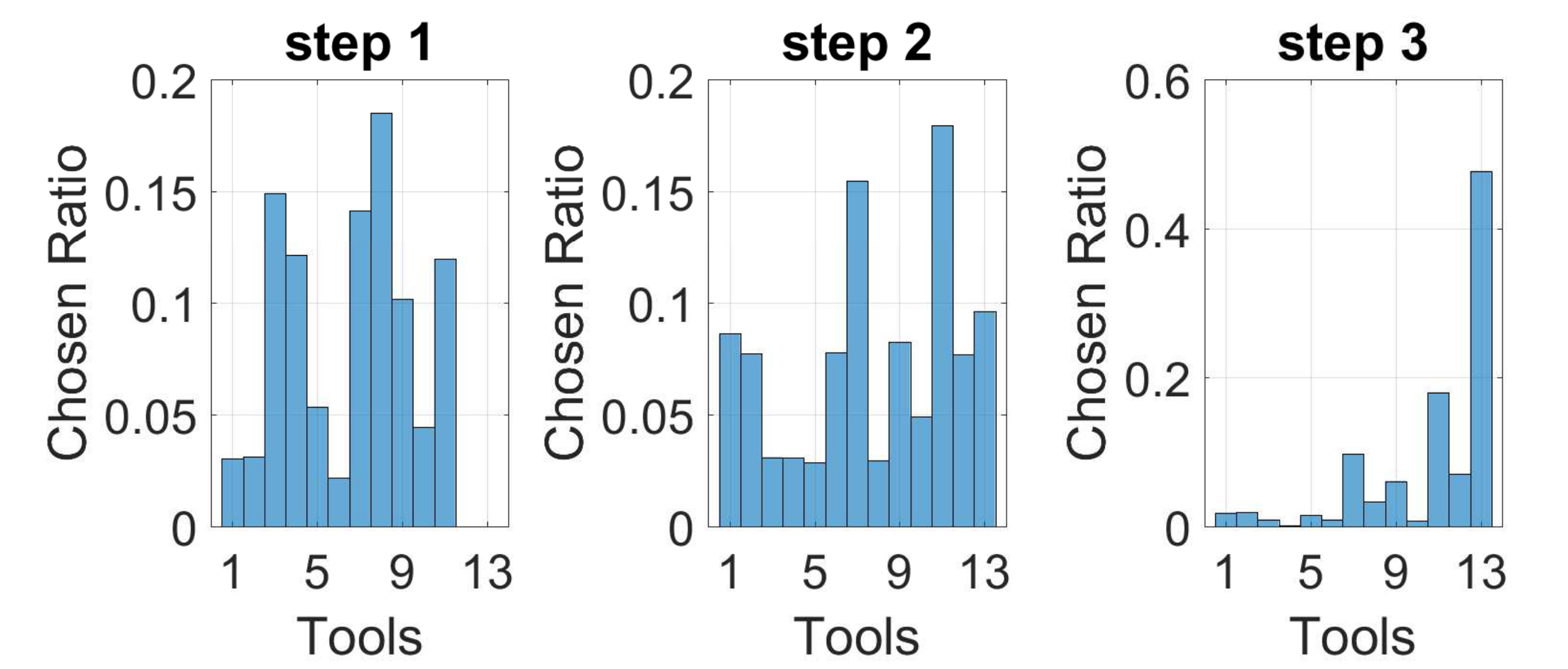}
		\caption{The chosen ratio of tool selection at each step.}
		\label{fig:actions}
	\end{center}
	\vspace{-0.75cm}
\end{figure}

To examine the internal behaviors of \textit{RL-Restore} , we analyze the frequency of the tool selection at each step. Results are shown in Figure~\ref{fig:actions}, where 0--12 on x-axis represent the 12 tools in Table~\ref{table:toolbox} and 13 is the stopping action. As can be observed on the three charts, the tool selection is diverse, and all tools are utilized in a different ratio. Specifically, deblurring and denoising tools are preferred at the first step, while denoising and de-JPEG tools are frequently chosen at the second step. The last step tends to stop the agent with a large probablity -- 47\%. Interestingly, when testing on unseen data, the ratios of stopping action at the last step are 60\% and 38\% on mild and severe test sets, respectively, which indicates that more severe and complex distortions require a longer toolchain to restore.


\begin{figure}[t]
	\begin{center}
		\includegraphics[width=\linewidth]{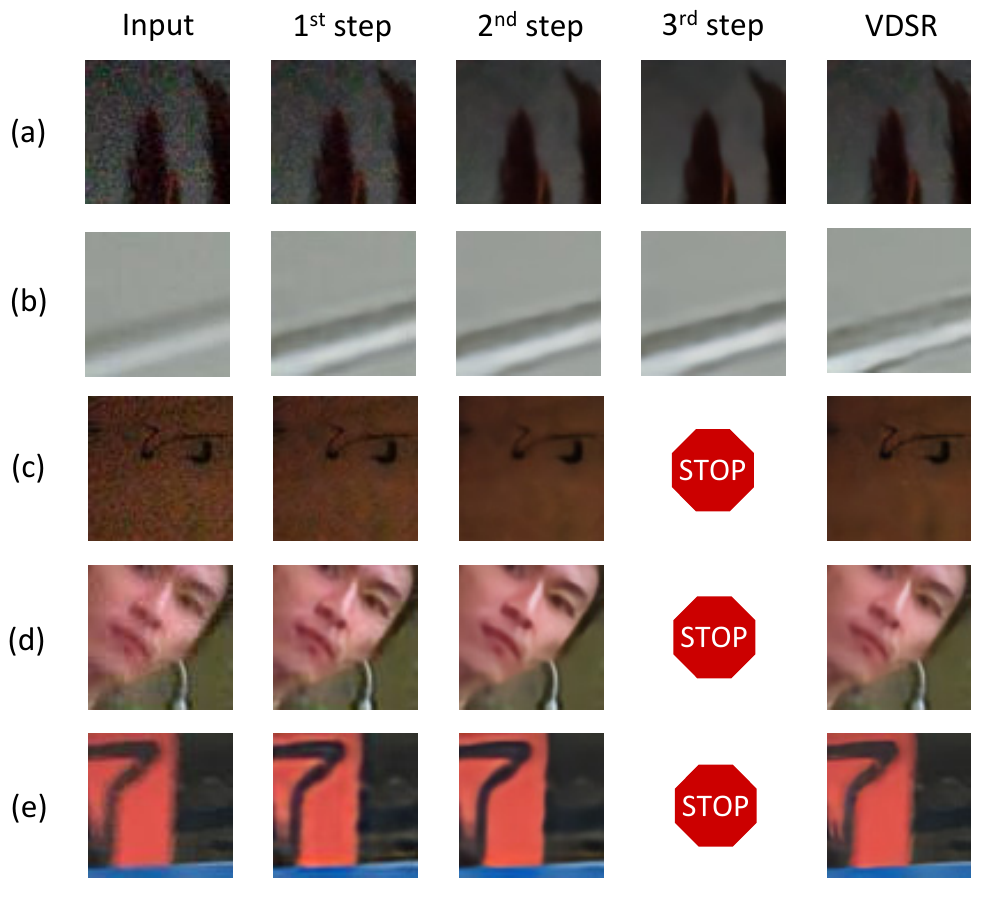}
		\caption{Results of real-world images.}
		\label{fig:real_world}
	\end{center}
	\vskip -0.6cm
\end{figure}

\subsection{Qualitative Evaluation on Real-World Images}
In real-world cases, images are always distorted by a variety of complex and mixed distortions with unknown degradation kernels, making restoration tasks extremely difficult for current methods. The proposed RL-based method may shed some light on possible solutions. When real-world distortions (\eg, slight out-of-focus blur, exposure noise and JPEG artifacts) are close to the training data, the proposed \textit{RL-Restore} can be easily generalized to these problems and performs better than a single CNN model. 

Examples are shown in Figure~\ref{fig:real_world}, where the input images, combined with different distortions (e.g., blurring, noise, compression), are captured by smart phones. We directly apply the well-trained \textit{RL-Restore} and VDSR on those real-world images, without further fine-tuning on the test data. It is obvious that our approach, benefiting from flexible toolchains, is more effective for restoring real-world images. Specifically, Figure~\ref{fig:real_world}(a, c) show that \textit{RL-Restore} can successfully deal with severe artifacts caused by exposure and compression, while Figure~\ref{fig:real_world}(b, d, e) demonstrate that our approach is able to restore a mix of blur and complex noise. It is also worth noting that the stopping action is selected by the agent when it is confident in the restored quality (Figure~\ref{fig:real_world}(c, d, e)). We believe that the proposed framework has the potential to deal with more complex real distortions with more powerful restoration tools.


\begin{table}[t] \centering \footnotesize
	\centering
	\caption{Ablation study on toolbox's size and toolchain's length.}
	\begin{tabular}{c|c|c|c|c|c|c|c}
		\hline
		\multicolumn{2}{c|}{Test Set}    &\multicolumn{2}{c|}{Mild (unseen)}  & \multicolumn{2}{c|}{Moderate} & \multicolumn{2}{c}{Severe (unseen)} \\ \hline
		\multicolumn{2}{c|}{Metric} & PSNR & SSIM & PSNR & SSIM & PSNR & SSIM \\
		\hline\hline
		\multirow{3}{*}{Size}  & 6 & 27.57 & 0.6241 & 25.72 & 0.5142 & 24.27 & 0.4291  \\ 
		& \hspace{-0.1cm}\underline{12} \hspace{-0.15cm} & \textbf{27.78} & \textbf{0.6372} & \textbf{26.20} & \textbf{0.5441} & \textbf{24.97} & 0.4643\\
		& \hspace{-0.1cm}18 \hspace{-0.15cm} & 27.77 & 0.6361 & 26.17 & 0.5417 & 24.93 & \textbf{0.4650} \\ 
		\hline
		\multirow{3}{*}{\hspace{-0.05cm}Length\hspace{-0.05cm}}  & 2 & 27.74 & 0.6264 & 25.99 & 0.5233 & 24.63 & 0.4444  \\ 
		& \underline{3} & \textbf{27.78} & \textbf{0.6372} & 26.20 & 0.5441 & 24.97 & 0.4643\\
		& 4 & 27.73 & 0.6368 & \textbf{26.20} & \textbf{0.5450} & \textbf{24.98} & \textbf{0.4663}\\ 
		\hline
	\end{tabular}
	\label{table:ablation_toolbox}
	\vskip -0.25cm
\end{table}

\subsection{Ablation Studies}

In this section, we investigate different settings of the proposed \textit{RL-Restore}, and give some insights on the choice of hyper-parameters. To better distinguish the effectiveness of each factor, we exclude the joint training strategy on all the experiments below. 

\noindent{\textbf{Toolbox Size and Toolchain Length.}}
The capacity of toolbox and the number of restoring actions dominate the restoration performance.
We alternatively vary the length of toolchain and the size of toolbox. 
As observed in Table~\ref{table:ablation_toolbox}, \textit{RL-Restore} performs well with $N=12$ and $T_{\mathrm{max}}=3$  under the current problem settings. 
Fewer tools and a shorter toolchain will decrease the performance. More tools and a longer toolchain achieve comparable performance. We attribute this phenomenon to the increased difficulty in learning more complex toolchains. It is worth pointing out that a toolchain with a length of two has a comparable PSNR as longer toolchains on the mild test set, indicating that slight distortions require fewer steps to restore.

\noindent{\textbf{Tools Training.}}
As discussed in Sec.~\ref{subsec:toolbox}, we propose two training strategies for tools to eliminate the complex artifacts in middle states: 1) Add slight noise and compression in the tools training data. 2) Perform joint training with the agent. Control experiments are conducted as in Table~\ref{table:ablation_tool_train}, where the `Original' setting represents the baseline, the `+Noise' adopts the first strategy and the `+Joint' uses both of them. It is obvious that adding noise to the training data successfully improves the PSNR by 0.2 dB, and joint training further pushes another 0.2 dB on all test sets, demonstrating the effectiveness of both training strategies.

\noindent{\textbf{Reward Function.}}
We experimentally find that the choice of reward functions can largely influence the performance. Besides the proposed stepwise reward based on PSNR, we also investigate other reward functions: 1) stepwise SSIM~\cite{wang2004image} where the reward is the SSIM gain at each step; 2) final PSNR where the reward is the final PSNR gain given at the last step; 3) final MSE as in~\cite{cao2017attention} where the reward is the negative MSE in the end. We adaptively adjust the learning rate for different rewards. As can be seen in Table~\ref{table:ablation_reward}, the stepwise SSIM, which performs the worst on PSNR metric, seems not to be a good choice for reward. The final MSE is slightly better on PSNR, but performs the worst on SSIM. The final PSNR achieves similar performance as the proposed stepwise PSNR reward. Nevertheless, we do not claim that PNSR is the best reward, and other evaluation methods are also encouraged for further comparison. 

\noindent{\textbf{Automatic Stopping.}}
The stopping action gives the agent the flexibility to terminate the restoration process when it is confident about the restored results. Thanks to this flexible stopping mechanism, it can prevent the images from over restored and save much computation. To demonstrate its effectiveness, we compare the results with/without the stopping action. As can be observed in Table~\ref{table:ablation_stopping}, the PSNR values drop around 0.15 dB when removing the stopping action. It is observed that the gap on mild test set is larger than that on other test sets. This is consistent with our experience that slight distortions are easily over restored if the agent does not stop in time.

\begin{table}[t] \centering \footnotesize
	\centering
	\caption{Ablation study on tools training.}
	\begin{tabular}{c|c|c|c|c|c|c}
		\hline
		Test Set  &\multicolumn{2}{c|}{Mild (unseen)}  & \multicolumn{2}{c|}{Moderate} & \multicolumn{2}{c}{Severe (unseen)} \\ \hline
		Metric & PSNR & SSIM & PSNR & SSIM & PSNR & SSIM \\
		\hline\hline
		+Joint  &  \textbf{28.04} & \textbf{0.6498} & \textbf{26.45} & \textbf{0.5587} & \textbf{25.20} & \textbf{0.4777} \\ 
		\hspace{0.15cm}+Noise \hspace{0.1cm}&  27.78 & 0.6372 & 26.20 & 0.5441 & 24.97 & 0.4643\\
		Original  & 27.52 & 0.6027 & 25.91 & 0.5119 & 24.81 & 0.4490\\ 
		\hline
	\end{tabular}
	\label{table:ablation_tool_train}
	\vskip -0.25cm
\end{table}

\begin{table}[t] \centering \footnotesize
	\centering
	\caption{Ablation study on reward functions.}
	\begin{tabular}{c|c|c|c|c|c|c}
		\hline
		Test Set  &\multicolumn{2}{c|}{Mild (unseen)}  & \multicolumn{2}{c|}{Moderate} & \multicolumn{2}{c}{Severe (unseen)} \\ \hline
		Metric & PSNR & SSIM & PSNR & SSIM & PSNR & SSIM \\
		\hline\hline
		\hspace{-0.1cm}\underline{Step. PSNR}\hspace{-0.1cm}  &  \textbf{27.78} & \textbf{0.6372} & \textbf{26.20} & \textbf{0.5441} & \textbf{24.97} & 0.4643 \\ 
		\hspace{-0.05cm}Step. SSIM\hspace{-0.1cm}& 26.58 & 0.6341 & 25.20 & 0.5368 & 24.18 & 0.4579 \\
		\hspace{-0.1cm}Final PSNR\hspace{-0.1cm}  & 27.71 & 0.6350 & 26.11 & 0.5417 & 24.86 & \textbf{0.4656}\\ 
		Final MSE & 27.14 & 0.6009 & 25.66 & 0.5166 & 24.55 & 0.4470\\
		\hline
	\end{tabular}
	\label{table:ablation_reward}
	\vskip -0.25cm
\end{table}

\begin{table}[t] \centering \footnotesize
	\centering
	\caption{Ablation study on stopping action.}
	\begin{tabular}{c|c|c|c|c|c|c}
		\hline
		Test Set  &\multicolumn{2}{c|}{Mild (unseen)}  & \multicolumn{2}{c|}{Moderate} & \multicolumn{2}{c}{Severe (unseen)} \\ \hline
		Metric & PSNR & SSIM & PSNR & SSIM & PSNR & SSIM \\
		\hline\hline
		\hspace{-0.15cm}w/ Stopping\hspace{-0.1cm}  &  \textbf{27.78} & \textbf{0.6372} & \textbf{26.20} & \textbf{0.5441} & \textbf{24.97} & \textbf{0.4643} \\ 
		\hspace{-0.15cm}w/o Stopping\hspace{-0.1cm} & 27.61 & 0.6284 & 26.08 & 0.5351 & 24.85 & 0.4589\\ \hline
	\end{tabular}
	\label{table:ablation_stopping}
	\vskip -0.25cm
\end{table}

\section{Conclusion}

We have presented a novel approach for image restoration based on reinforcement learning. Unlike most existing deep learning based methods, in our approach an agent is learned to dynamically select a toolchain to progressively restore an image that is corrupted by complex and mixed distortions. Extensive results on synthetic and real-world images validate the effectiveness of the proposed approach. With its inherent flexibility, the proposed framework can be applied to more challenging restoration tasks or other low-level vision problems by developing powerful tools and an appropriate reward.

\vspace{0.1cm}
\noindent
\textbf{Acknowledgement}. This work is supported by SenseTime Group Limited and the General Research Fund sponsored by the Research Grants Council of the Hong Kong SAR (CUHK 14241716, 14224316. 14209217).

{\small
	\bibliographystyle{ieee}
	\bibliography{bib_v1}
}

\end{document}